\documentclass[10pt,twocolumn,letterpaper]{article}

\usepackage{wacv}
\usepackage{times}
\usepackage{epsfig}
\usepackage{graphicx}
\usepackage{amsmath}
\usepackage{amssymb}

\usepackage{algorithm}
\usepackage{algpseudocode}
\usepackage{amsmath,amssymb}
\usepackage{sidecap}
\usepackage{caption}
\usepackage{enumitem}
\usepackage{subfigure}
\DeclareMathOperator{\E}{\mathbb{E}}

\usepackage{tabularx}
\newcolumntype{b}{>{\centering\arraybackslash}X}
\newcolumntype{d}{>{\raggedleft\arraybackslash}X}
\newcolumntype{s}{>{\hsize=.5\hsize}X}
\newcolumntype{m}{>{\hsize=.35\hsize}b}
\newcolumntype{z}{>{\hsize=.75\hsize}b}

\wacvfinalcopy 

\def\wacvPaperID{134} 

\ifwacvfinal\fi
\begin{document}

\title{ Local Gradients Smoothing: Defense against localized adversarial attacks}

\author{Muzammal Naseer \\
 Australian National University (ANU)\\
{\tt\small muzammal.naseer@anu.edu.au}
\and
Salman H. Khan \\
Data61, CSIRO\\
{\tt\small salman.khan@data61.csiro.au}
\and
Fatih Porikli \\
 Australian National University (ANU)\\
{\tt\small fatih.porikli@anu.edu.au}
}

\maketitle
\ifwacvfinal\thispagestyle{empty}\fi

\begin{abstract}
  Deep neural networks (DNNs) have shown vulnerability to adversarial attacks, i.e., carefully perturbed inputs designed to mislead the network at inference time. Recently introduced localized attacks, Localized and Visible Adversarial Noise (LaVAN) and Adversarial patch, pose a new challenge to deep learning security by adding adversarial noise only within a specific region without affecting the salient objects in an image. Driven by the observation that such attacks introduce concentrated high-frequency changes at a particular image location, we have developed an effective method to estimate noise location in gradient domain and transform those high activation regions caused by adversarial noise in image domain while having minimal effect on the salient object that is important for correct classification. Our proposed Local Gradients Smoothing (LGS) scheme achieves this by regularizing gradients in the estimated noisy region before feeding the image to DNN for inference. We have shown the effectiveness of our method in comparison to other defense methods including Digital Watermarking, JPEG compression,  Total Variance Minimization (TVM)  and Feature squeezing on ImageNet dataset. In addition, we systematically study the robustness of the proposed defense mechanism against Back Pass Differentiable Approximation (BPDA), a state of the art attack recently developed to break defenses that transform an input sample to minimize the adversarial effect. Compared to other defense mechanisms, LGS is by far the most resistant to BPDA in localized adversarial attack setting.
\end{abstract}

\section{Introduction}

\label{sec:intro}
Deep neural network architectures achieve remarkable performance on critical applications of machine learning including sensitive areas such as face detection \cite{parkhi2015deep}, malware detection \cite{ronen2018microsoft} and autonomous driving \cite{huval2015empirical}. However, the vulnerability of DNNs to adversarial examples limit their wide adoption in security critical applications \cite{Akhtar2018ThreatOA}. It has been shown that adversarial examples can be created by minimally modifying the original input samples such that a DNN mis-classifies them with high confidence. DNNs are often criticized as black-box models; adversarial examples raise further concerns by highlighting blind spots of DNNs. At the same time, adversarial phenomena provide an opportunity to understand DNN's behavior to minor perturbations in visual inputs. 

\begin{figure*}
\centering
\begin{minipage}[t]{0.7\linewidth}
  \subfigure[  Impala (94\%)]{\includegraphics[width=0.32\textwidth, height=4cm]{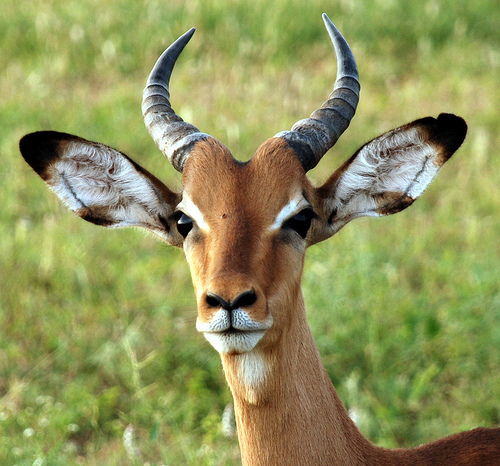}}
  \subfigure[  Ice Lolly (99\%)]{\includegraphics[width=0.32\textwidth, height=4cm]{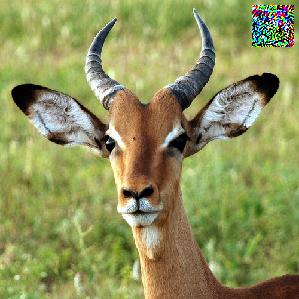}}
  \subfigure[  Impala (94\%)]{\includegraphics[width=0.32\textwidth, height=4cm]{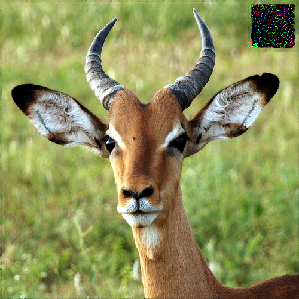}}
  
  \subfigure[  Squirrel Monkey (58\%)]{\includegraphics[width=0.32\textwidth, height=4cm]{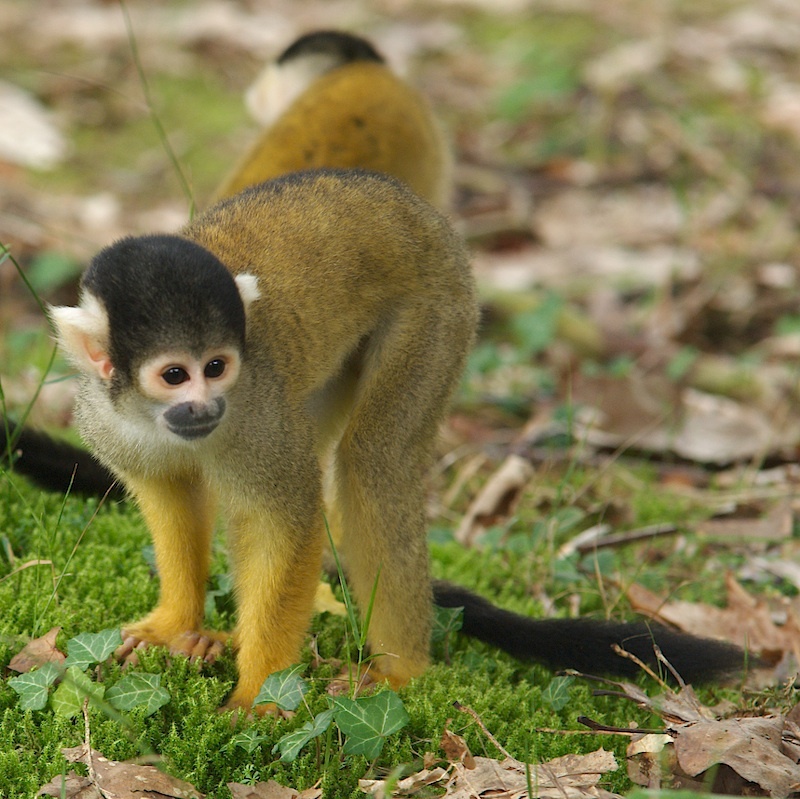}}
  \subfigure[  Toaster (91\%)]{\includegraphics[width=0.32\textwidth, height=4cm]{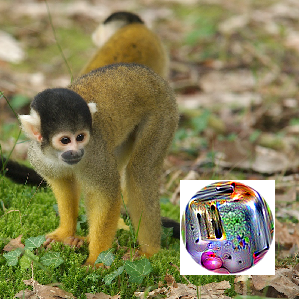}}
  \subfigure[Squirrel Monkey (57\%)]{\includegraphics[width=0.32\textwidth, height=4cm]{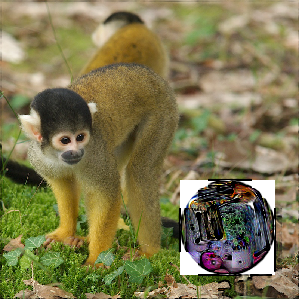}} 
\end{minipage}
\captionof{figure}{Inception v3 \cite{szegedy2016rethinking} confidence scores are shown for example images. (a) and (d) represent benign examples from ImageNet  \cite{ILSVRC15}, (b) and (e) are adversarial examples generated by LaVAN \cite{karmon2018lavan} and Adversarial patch \cite{brown2017adversarial} respectively, (c) and (f) show transformed adversarial images using our proposed LGS. As illustrated, LGS restores correct class confidences.}
\label{fig:intro_fig}
\end{figure*}

Methods that generate adversarial examples either modify each image pixel by a small amount \cite{szegedy2013intriguing,goodfellow2014explaining,moosavi2016deepfool,madry2017towards} often imperceptible to human vision or few image pixels by a large visible amounts \cite{sharif2016accessorize,su2017one, brown2017adversarial, karmon2018lavan, evtimov2017robust}. Pixel attack \cite{su2017one} changes few image pixels, but it requires small images (e.g., 32$\times$32) and does not provide control over noise location. Small noise patches were introduced by \cite{sharif2016accessorize} in the form of glasses to cover human face to deceive face recognition systems. Similarly, Evtimov \etal \cite{evtimov2017robust} added noise patches as rectangular patterns on top of traffic signs to cause misclassification. Very recently, localized adversarial attacks, i.e., Adversarial patch \cite{brown2017adversarial} and  LaVAN \cite{karmon2018lavan} have been introduced that can be optimized for triplets (misclassification confidence, target class, perturbed location). These practical attacks have demonstrated high strength and can easily bypass existing defense approaches. Therefore they present a significant challenge for existing deep learning systems.

\textbf{Contributions:} In this work, we study the behavior of localized adversarial attacks and propose an effective mechanism to defend against them (see Fig.~\ref{fig:intro_fig}). LaVAN and Adversarial patch add adversarial noise without affecting the original object in the image, and to some extent, they are complementary to each other. In an effort towards a strong defense against these attacks, this paper contributes as follows:
\begin{itemize}\setlength{\itemsep}{0em}
\item Motivated by the observation that localized adversarial attacks introduce high-frequency noise, we propose a transformation called Local Gradient Smoothing (LGS). LGS first estimates region of interest in an image with the highest probability of adversarial noise and then performs gradient smoothing in only those regions.
\item We show that by its design, LGS significantly reduces gradient activity in the targeted attack region and thereby showing the most resistance to BPDA \cite{athalye2018obfuscated}, an attack specifically designed to bypass transformation based defense mechanisms.
\item Our proposed defense outperforms other state-of-the-art methods such as Digital watermarking, TVM, JPEG compression, and Feature squeezing in localized adversarial attacks setting \cite{karmon2018lavan,brown2017adversarial}.
\end{itemize}

\section{Related Work}
Among the recent localized adversarial attacks, the focus of adversarial patch \cite{brown2017adversarial} is to create a scene independent physical-world attack that is agnostic to camera angles, lighting conditions and even the type of classifier. The result is an image independent universal noise patch that can be printed and placed in the classifier's field of view in a white box (when deep network model is known) or black box (when deep network model is unknown) setting. However, the size of the adversarial patch should be 10\% of the image for the attack to be successful in about 90\% cases \cite{karmon2018lavan}. This limitation was addressed by Karmoon \etal \cite{karmon2018lavan}, who focused on creating localized attack covering as little as 2\% of the image area instead of generating  a universal noise patch. In both of these attacks \cite{brown2017adversarial,karmon2018lavan}, there is no constraint on noise, and it can take any value within image domain, i.e., [0, 255] or [0, 1].

Defense mechanisms against adversarial attacks can be divided into two main categories. (a) Methods that modify DNN by using adversarial training \cite{tramer2017ensemble} or gradient masking \cite{papernot2016distillation} and (b) techniques that modify input sample by using some smoothing function to reduce adversarial effect without changing the DNN \cite{dziugaite2016study, das2018shield, guo2017countering, xu2017feature}. For example, JPEG compression was first presented as a defense by \cite{dziugaite2016study} and recently studied extensively by \cite{das2018shield,shaham2018defending}. \cite{xu2017feature} presented feature squeezing methods including bit depth reduction, median filtering, Gaussian filtering to detect and defend against adversarial attacks. Guo \etal \cite{guo2017countering} considered smoothing input samples by total variance minimization along with JPEG compression and image quilting to reduce the adversarial effect. Our work falls into the second category as we also transform the input sample to defend against localized adversarial attacks. However, as we will demonstrate through experiments, the proposed defense mechanism provides better defense against localized attacks compared to previous techniques. 

The paper is organized as follows: Section~\ref{sec:attacks} discusses localized adversarial attacks, LaVAN and Adversarial patch in detail. Section~\ref{sec:LGS} presents our defense approach (LGS) against these attacks. We discuss other related defense methods in Section~\ref{sec: RW}. Section~\ref{sec:Expts} demonstrates the effectiveness of the proposed method LGS in comparison to other defense methods against LaVAN and adversarial patch attacks. Section ~\ref{sec: BPDA} discusses BPDA and resilience of different defense methods against it. Section ~\ref{sec: Conclu} concludes the draft by discussing possible future directions.

\section{Adversarial Attacks}
\label{sec:attacks}
In this section, we provide a brief background to adversarial attacks and explain how LaVAN \cite{karmon2018lavan} and Adversarial patch \cite{brown2017adversarial} are different from traditional attacks.

\subsection{Traditional Attacks} 
The search for adversarial examples can be formulated as a constrained optimization problem. Given a discriminative classifier $\mathcal{F}(\mathbf{y}\,|\,\mathbf{x})$, an input sample $\mathbf{x}\, \in \, \mathbb{R}^{n}$, a target class $\bar{\mathbf{y}}$ and a perturbation budget $\mathbf{\epsilon}$, an attacker seeks to find a modified input $\mathbf{x}'=\mathbf{x}+\mathbf{\delta} \in \, \mathbb{R}^{n}$ with adversarial noise $\delta$ to increase likelihood of the target class $\bar{\mathbf{y}}$ by solving the following optimization problem: 
\begin{align}\label{eq: adv_att}
 \underset{\mathbf{\mathbf{x}'}}{\text{max}} & \;\; \mathcal{F}(\mathbf{y} = \bar{\mathbf{y}}\,|\, \mathbf{x}' ) \notag \\
 \text{subject to:} & \;\;  \Vert \mathbf{x} - \mathbf{x}' \Vert_{p} \leq \epsilon
\end{align}
This formulation produces well camouflaged adversarial examples but changes each pixel in the image. Defense methods such as JPEG compression \cite{dziugaite2016study,das2018shield}, Total variance minimization \cite{guo2017countering} and Feature squeezing \cite{xu2017feature} are effective against such attacks especially when the perturbation budget $\mathbf{\epsilon}$ is not too high.

\subsection{LaVAN}
\label{subsec:lavan}
LaVAN \cite{karmon2018lavan} differs from the formulation presented in Eq.~\ref{eq: adv_att} as it confines adversarial noise $\delta$ to a small region, usually away from the salient object in an image. It uses the following spatial mask to replace the small area with noise, as opposed to noise addition performed in traditional attacks: 
\begin{equation}
\label{eq: mask}
\mathbf{x}' = (\mathbf{1}-\mathbf{m})\odot \mathbf{x}+\mathbf{m}\odot \mathbf{\delta},  \quad s.t., \mathbf{m} \in \mathbb{R}^n  \text{ and },
\end{equation} 
where $\odot \hspace{0.1cm} \text{is Hadamard product}$ and $\delta$ represents adversarial noise. 

They also introduce a new objective function where at each iteration, optimization algorithm takes a step away from the source class and towards the target class simultaneously, as follows:
\begin{align}
 \underset{\mathbf{x}'}{\text{max}} & \;\; \mathcal{F}(\bar{\mathbf{y}}\,|\, \mathbf{x}' ) \notag -  \mathcal{F}(\mathbf{y}\,|\, \mathbf{x}' ) \notag  \\
 \text{subject to:} & \;\; \Vert \mathbf{x} - \mathbf{x}' \Vert_{\infty} \leq \mathbf{\epsilon}, \;\; 0 \leq \epsilon \leq 1,
\end{align}
where $\mathbf{x'}$ is given by Eq.~\ref{eq: mask}. 

\subsection{Adversarial Patch}
\label{subsec: adv_p}
Adversarial examples created using the methodology presented in Eq.~\ref{eq: adv_att} cannot be used in physical world attacks because adversarial noise loses its effect under different camera angles, rotations and lighting conditions. Athalye \etal \cite{athalye2017synthesizing} introduced an Expectation over Transformation (EoT) attack to create robust adversarial examples invariant to chosen set of transformations. Brown \etal \cite{brown2017adversarial} build upon Athalye's work and used EoT to create a scene independent robust noise patch confined to small region that can be printed and placed in the classifier's field of view to cause misclassification. To generate adversarial patch $\mathbf{p}'$, \cite{brown2017adversarial} proposed a patch operator $\mathcal{A}(\mathbf{p}, \mathbf{x}, \mathbf{l}, \mathbf{t})$ for a given image $\mathbf{x}$, patch $\mathbf{p}$, location $\mathbf{l}$ and a set of transformation $\mathbf{t}$. During optimization, patch operator $\mathcal{A}$ applies a set of transformations to the patch $\mathbf{p}$ and then projects it onto the image $\mathbf{x}$ at a location $\mathbf{l}$ to increase likelihood of target class $\bar{\mathbf{y}}$. 
\begin{equation} 
\label{eq:patch}
    \mathbf{p}' =\underset{\mathbf{p}}{\text{max}} \E_{\mathbf{x} \sim {X}, \mathbf{t} \sim {T}, \mathbf{l} \sim {L}}[{\mathcal{F}(\bar{\mathbf{y}}\,|\, \mathcal{A}(\mathbf{p}, \mathbf{x}, \mathbf{l}, \mathbf{t}))}]
\end{equation}
where $X$ represent training images, $T$ represents distribution over transformations, and $L$ is a distribution over locations in the image.

\begin{figure*}
\centering
\noindent\begin{minipage}{\textwidth}
  \centering
  \begin{minipage}{.20\textwidth}
  	\centering
    \includegraphics[width=\linewidth,  height=3.5cm]{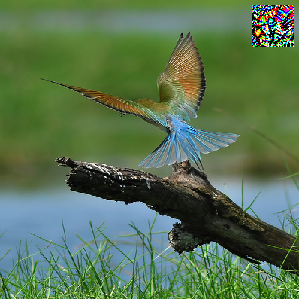}
    \scriptsize (a)
  \end{minipage}
   \begin{minipage}{.20\textwidth}
   	\centering
    \includegraphics[width=\linewidth, height=3.5cm]{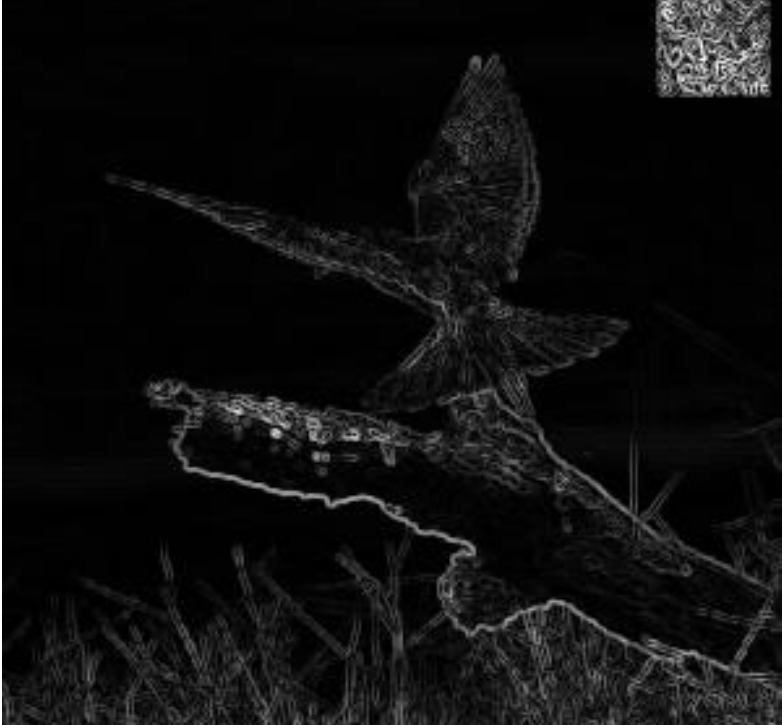}\
     \scriptsize (b)
  \end{minipage}
  \begin{minipage}{.20\textwidth}
  	\centering
    \includegraphics[width=\linewidth,  height=3.5cm]{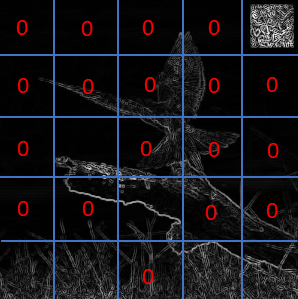}\
    \scriptsize (c)
  \end{minipage}
  \begin{minipage}{.20\textwidth}
  	\centering
     \includegraphics[width=\linewidth,  height=3.5cm]{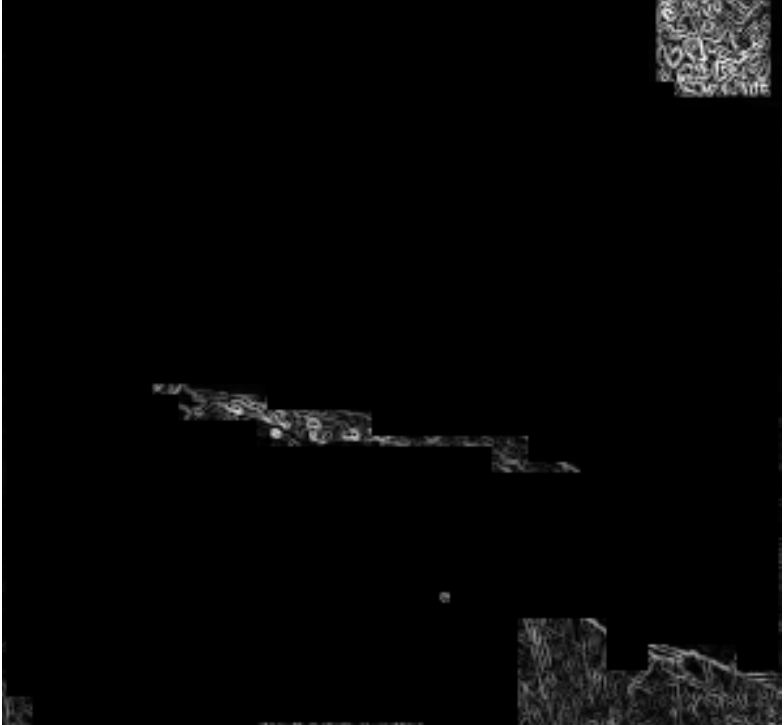}\
    \scriptsize (d)
  \end{minipage}
 \vspace{0.01\textwidth}
\end{minipage}
\noindent\begin{minipage}{\textwidth}
  \centering
  \begin{minipage}{.20\textwidth}
  	\centering
    \includegraphics[width=\linewidth,  height=3.5cm]{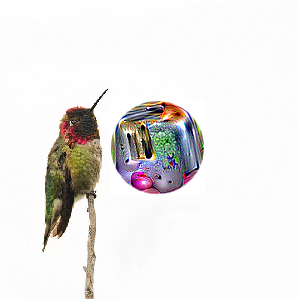}\
    \scriptsize (e) 
  \end{minipage}
   \begin{minipage}{.20\textwidth}
   	\centering
    \includegraphics[width=\linewidth,  height=3.5cm]{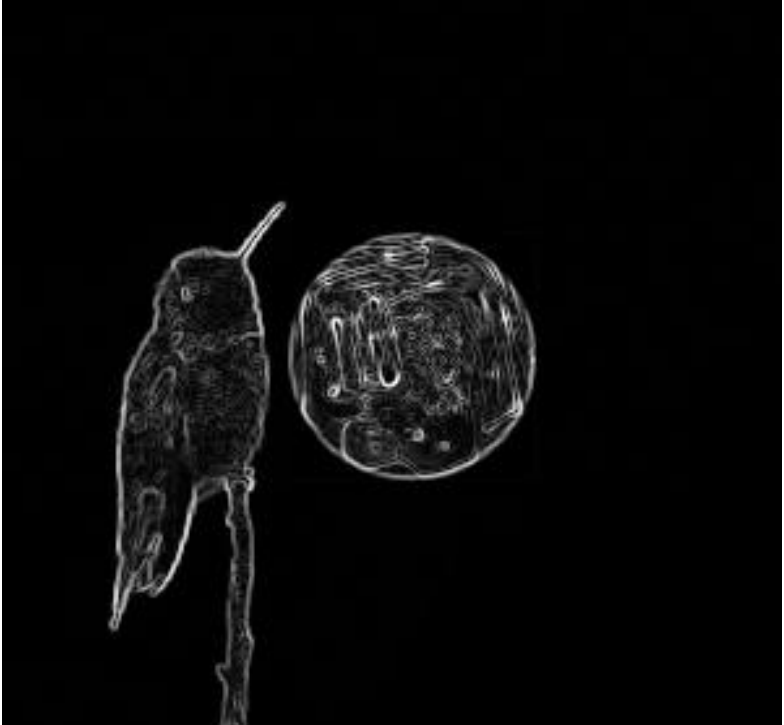}\
     \scriptsize (f) 
  \end{minipage}
  \begin{minipage}{.20\textwidth}
  	\centering
    \includegraphics[width=\linewidth,  height=3.5cm]{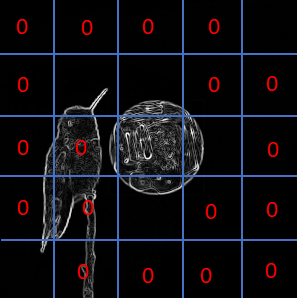}\
    \scriptsize (g)
  \end{minipage}
  \begin{minipage}{.20\textwidth}
  	\centering
   \includegraphics[width=\linewidth,  height=3.5cm]{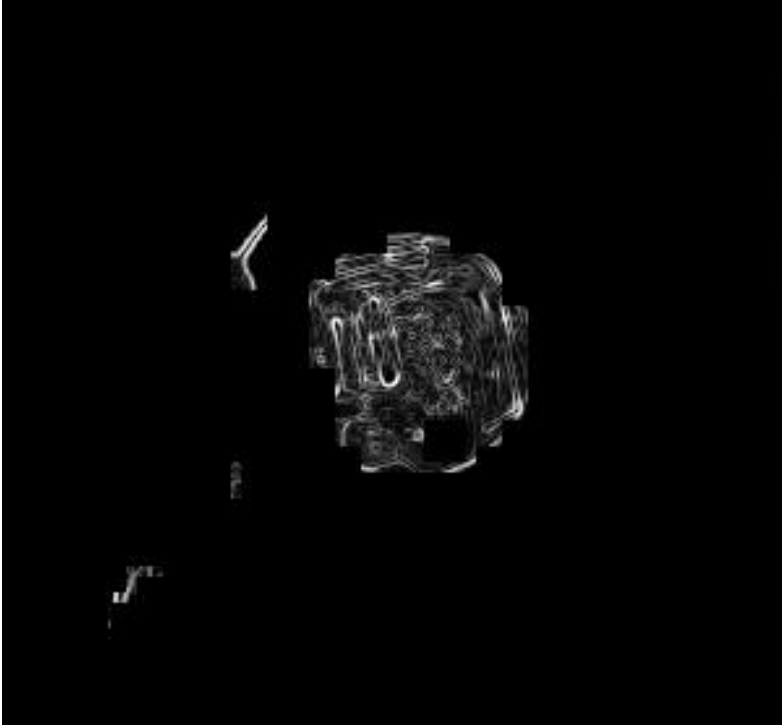}\
    \scriptsize (h)
  \end{minipage}
    \vspace{0.01\textwidth}
\end{minipage}
\vspace{0.1cm}
\caption{(a) and (e) show adversarial examples generated by LaVAN \cite{karmon2018lavan} and adversarial patch \cite{brown2017adversarial} respectively, (b) and (f) show normalized gradients magnitude before applying windowing operation to look for highest activation regions, (c) and (g) show concept of window search to estimate noise regions, (d) and (h) show normalized gradients magnitude after applying windowing operation. We refer to supplementary material for more examples.}
\label{fig:lgs_smoothing}
\end{figure*}

\section{Defense: Local Gradients Smoothing}
\label{sec:LGS}
Both of the above discussed attacks \cite{karmon2018lavan,brown2017adversarial} introduce high frequency noise concentrated at a particular image location and strength of such a noise becomes very prominent in image gradient domain. We propose that the effect of such adversarial noise can be reduced significantly by suppressing  high frequency regions without effecting the low frequency image areas that are important for classification. An efficient way to achieve this is by projecting scaled normalized gradient magnitude map onto the image to directly suppress high activation regions. To this end, we first compute the magnitude of first-order local image gradients as follows: 
\begin{equation}\label{eq:local_grad}
   \parallel\nabla \mathbf{x}(a,b)\parallel = \sqrt{\left (  \frac{\partial \mathbf{x}}{\partial a}\right )^{2} + \left (  \frac{\partial \mathbf{x}}{\partial b}\right )^{2}},
\end{equation}
where $a, b$ denote the horizontal and vertical directions in the image plane. The range of the gradient magnitude calculated using the above equation is normalized for consistency across an image as follows:
\begin{equation}
g(\mathbf{x}) = \frac{\parallel\nabla \mathbf{x}(a,b)\parallel - \parallel\nabla \mathbf{x}(a,b)\parallel_{min}}{\parallel\nabla \mathbf{x}(a,b)\parallel_{max} - \parallel\nabla \mathbf{x}(a,b)\parallel_{min}}.
\end{equation}
The normalized gradient $g(\mathbf{x})$ is projected onto the original image to suppress noisy perturbations in the input data domain. This operation smooths out very high frequency image details. As demonstrated by our evaluations, these regions have high likelihood of being perturbed areas, but they do not provide significant information for final classification. The noise suppression is performed as follows: 
\begin{equation}\label{eq:proj}
\mathcal{T}(\mathbf{x})= \mathbf{x}\odot(1-\lambda*g(\mathbf{x})),
\end{equation}
where $\lambda$ is the smoothing factor for LGS and $\lambda*g(x)$ is clipped between 0 and 1. Applying this operation at a global image level, however, introduces image structural loss that causes a  drop in classifier's accuracy on benign examples. To minimize this effect, we design a block-wise approach where gradient intensity is evaluated within a local window. To this end, we first divide the gradient magnitude map into a total of $K$ overlapping blocks of same size ($\tau$) and then filter these blocks based on a threshold ($\gamma$) to estimate highest activation regions which also have the highest likelihood of adversarial noise. This step can be represented as follows:
\begin{align}
\label{eq:wind}
\mathbf{g}'_{h,w} &= \mathcal{W}(g(\mathbf{x}),h,w,\tau, o) \in \mathbb{R}^{\tau}, \notag \\
  \hat{\mathbf{g}}_{h,w} &= \begin{cases}
    \mathbf{g}'_{h,w}, & \text{if $\frac{1}{|\mathbf{g}'_{h,w}|}\sum_i\sum_j {g}'_{h,w}(i,j) > \gamma$}\\
    0, & \text{otherwise}.
  \end{cases}    
\end{align}
where $|\cdot|$ denotes the cardinality of each patch, $o$ denotes the patch overlap, $\mathcal{W}(\cdot)$ represent the windowing operation, $h,w$ denote the vertical and horizontal components of the top left corner of the extracted window, respectively. We set the block size $\tau = 15\times 15$ with $5\times 5$ overlap and threshold is $0.1$ in all of our experiments. The updated gradient blocks represented as $\hat{\mathbf{g}}_{h,w}$ are then collated to recreate the full gradient image: $\bar{\mathbf{g}} = \mathcal{W}^{-1}(\{\hat{\mathbf{g}}_{h,w}\}_{1}^{K})$. Figure~\ref{fig:lgs_smoothing} shows the effect of windowing search on gradients magnitude maps. We further demonstrated LGS efficiency on challenging images in supplementary material.

\begin{table*}[h]
\centering
\scalebox{0.85}{
\begin{tabularx}{.95\textwidth}{p{3cm}>{\centering\arraybackslash}p{1.5cm}>{\centering\arraybackslash}p{3.5cm}>{\centering\arraybackslash}p{3.5cm}>{\centering\arraybackslash}p{3.5cm}}
\hline
          & No Attack & 42x42 noise patch covering $\sim$2\% of image & 52x52 noise patch covering $\sim$3\% of image & 60x60 noise patch covering $\sim$ 4\% of image \\
      \hline
      No Defense & 75.61\% & 11.00\% & 2.79\% & 0.78\%\\ 
      \hline
      LGS [lambda=2.3] & 71.05\% & \bf{70.90\%} & \bf{69.84\%} & \bf{69.37\%}\\
      LGS [lambda=2.1] & 71.50\%  & 70.80\%  & 69.54\% & 68.56\%\\
      LGS [lambda=1.9] & 71.84\%  & 70.40\% & 68.84\% & 66.98\%\\
      LGS [lambda=1.7] & 72.30\% & 69.55\% & 67.32\% & 63.38\%\\
      LGS [lambda=1.5] & 72.72\% & 67.68\% & 64.13\% & 55.67\%\\
      \hline
      DW              & 52.77\% & \bf{67.70\%} & \bf{66.19\%} & \bf{64.57\%}\\
      \hline
      MF [window=3] & 70.59\% & \bf{63.90\%} & \bf{62.15\%}& \bf{59.81}\% \\
      GF [window=5] & 61.75\% & 59.52\%  & 57.68\% & 55.29\%\\
      BF [window=5] & 65.70\% & 61.53\% & 58.70\% & 55.59\%\\
      \hline
     JPEG [quality=80]  & 74.35\% & 18.14\% & 6.23\% & 2.06\%\\
     JPEG [quality=60]  & 72.71\% & 25.69\% & 11.86\% & 4.85\%\\
     JPEG [quality=40]  & 71.20\% & 37.10\% & 23.26\% & 12.73\%\\
     JPEG [quality=30]  & 70.04\% & 45.00\% & 33.72\% & 22.04\%\\
     JPEG [quality=20]  & 67.51\% & 52.84\%  & 46.25\% & 37.19\% \\
     JPEG [quality=10]  & 60.25\% & \bf{53.10\%} & \bf{48.73\%} & \bf{43.59\%}\\
     \hline
     TMV [weights=10] & 70.21\% & \bf{14.48\%} & \bf{4.64\%} & \bf{1.73\%}\\
     TMV [weights=20] & 72.85\% & 13.24\% & 3.78\% & 1.17\%\\
     TMV [weights=30] & 73.85\% & 12.79\% & 3.53\% & 1.04\%\\
     \hline
     BR [depth=1] & 39.85\% & \bf{25.93\%} & \bf{15.14\%} & \bf{9.73\%}\\
     BR [depth=2] & 64.61\% & 16.32\% & 6.15\% & 2.68\%\\
     BR [depth=3] & 72.83\% & 13.4\% & 3.89\% & 1.25\%\\
     \hline
    \end{tabularx}}
    \vspace{0.2cm}
    \caption{Summary of Inception v3 performance against LaVAN attack on ImageNet validation set with and without defenses including local gradient smoothing (LGS), digital watermarking (DW), median filtering (MF), Gaussian filtering (GF), bilateral filtering (BF), JPEG compression, total variance minimization (TVM) and bit-depth reduction (BR). Bold numbers represent the best accuracy of a certain defense against LAVAN attack.}
    \label{tab:accuracy}
\end{table*}

\section{Experiments}
\label{sec:Expts}

\subsection{Protocol and Results Overview}
We used Inception v3 model \cite{szegedy2016rethinking} to experiment with various attack and defense mechanisms in all of our experiments. All attacks are carried out in white-box settings. We consider the validation set available with Imagenet-2012 dataset in our experiments. This set consists of a total of 50k images. We report top-1 accuracy of classifier. Results are summarized in tables ~\ref{tab:accuracy}, ~\ref{tab:adv_p} and \ref{tab:BPDA}. 

\begin{figure}[h]
    \centering
    \includegraphics[width=\linewidth, height=4.5cm]{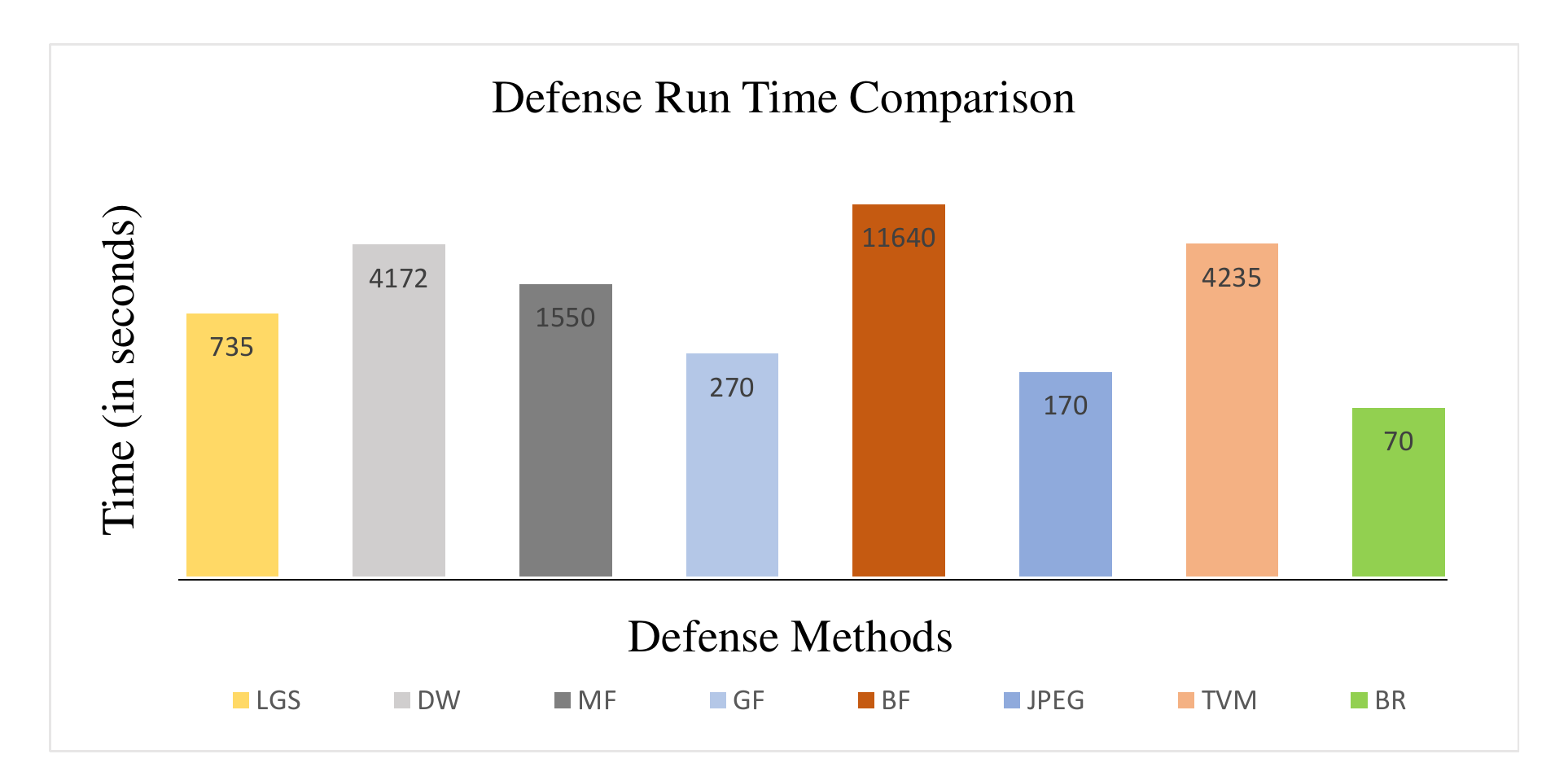}
    \caption{Computational cost comparison of defense methods to process 50k ImageNet validation images. Graph is shown in log scale for better visualization with actual processing times written on the top of each bar in seconds.}
    \label{fig: compare}
\end{figure}

LaVAN \cite{karmon2018lavan} can be optimized for triplets (target, confidence, location) but it is highly sensitive to noise location. Adversary loses its effect with even a small change to the pixel location. To reduce the computational burden and conduct experiments on a large scale, we randomly chose noise location along border areas of the image because they have the least probability to cover the salient object. We ran 1000 iterations of attack optimization per image. We terminate the optimization early if classifier mis-classify with confidence above than or equal to 99\% or we let it run for at max 1000 iterations and attack is considered to be successful if the image label is changed to a random target (not equal to the true object class). Inceptionv3 model accepts 299x299 image as an input. Three adversarial noise masks with size 42x42 ($\sim$2\% of the image),  52x52 ($\sim$3\% of the image) and 60x60 ($\sim$4\% of the image) were applied. Table~\ref{tab:accuracy} presents summary of all the results. For the case of adversarial patch \cite{brown2017adversarial} attack, placing a patch of size 95x95 (~10\% of the image) randomly on all Imagenet validation set was not possible because it would cover most of salient objects details in an image. So we carefully created 1000 adversarial examples that model misclassified as a toaster with a confidence score at least 90\%. We then applied all the defense techniques and reported results in Table~\ref{tab:adv_p}. Figure ~\ref{fig: compare} shows runtime of defense methods to process ImageNet \cite{ILSVRC15} validation set. We used optimized python implementations. Specifically, we employed JPEG from Pillow, Total variance minimization (TVM), and Bilateral filtering (BF) from scikit-image, Median filtering (MF) and Gaussian filtering (GF) from scipy, and LGS and Bit Depth Reduction (BR) are written in python 3.6 as well. All experiments were conducted on desktop windows computer equipped with Intel i7-7700k quad-core CPU clocked at 4.20GHz and 32GB RAM.

\begin{table}[h]
\centering
\scalebox{0.8}{
\begin{tabularx}{0.6\textwidth}{@{}p{2cm}ccccccc@{}} 
\hline
Defense   & None     &  LGS & DW & MF & JPEG & TVM & BR \\ \hline
Adversarial Patch   & 0\%  & \bf{90.5\%} & 80\%&  49.10\%   & 45\%   &  1\% &   0\%   \\ \hline
\end{tabularx}}
\vspace{0.2cm}
\caption{Accuracy of Inception v3 against adversarial patch attack with and without defense. The size of adversarial noise is 95x95 covering $\sim$10\% of image. LGS is used with $\lambda=2.3$, DW in blind defense scenario, MF with window equal to 3, JPEG compression with quality equal to 30, TVM with weights equal to 10 and BR with depth 3. This hyperparameter choice was made for fair comparison such that the performance on benign examples from ImageNet is approximately the same (first column of Table \ref{tab:accuracy}). Results are reported for 1000 adversarial examples misclassified as toaster with confidence above than 90\%.}
\label{tab:adv_p}
\end{table}

\begin{figure*}
\centering
\noindent\begin{minipage}{\textwidth}
  \centering  
  \begin{minipage}{.135\textwidth}
  	\centering
    \includegraphics[width=\linewidth, keepaspectratio]{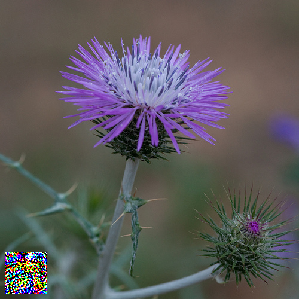}\
    \scriptsize (a) Dragonfly \\ (99\%)
  \end{minipage}
   \begin{minipage}{.135\textwidth}
   	\centering
    \includegraphics[width=\linewidth,keepaspectratio]{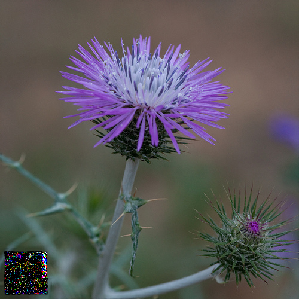}\
     \scriptsize (c) Cardoon \\ (94\%)
  \end{minipage}
  \begin{minipage}{.135\textwidth}
  	\centering
    \includegraphics[width=\linewidth, keepaspectratio]{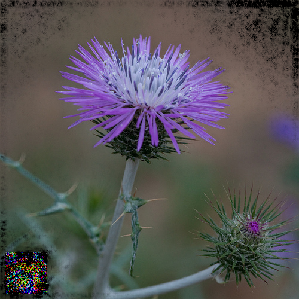}\
    \scriptsize (e) Cardoon \\ (91\%)
  \end{minipage}
  \begin{minipage}{.135\textwidth}
  	\centering
     \includegraphics[width=\linewidth, keepaspectratio]{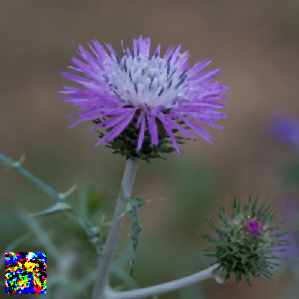}\
    \scriptsize (g) Cardoon \\ (89\%)
  \end{minipage}
    \begin{minipage}{.135\textwidth}
  	\centering
     \includegraphics[width=\linewidth, keepaspectratio]{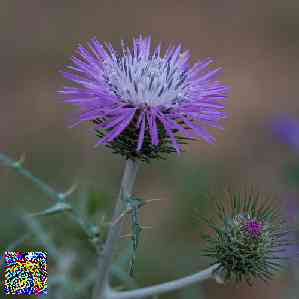}\
    \scriptsize (i) Dragonfly \\ (70\%)
  \end{minipage}
    \begin{minipage}{.135\textwidth}
  	\centering
     \includegraphics[width=\linewidth, keepaspectratio]{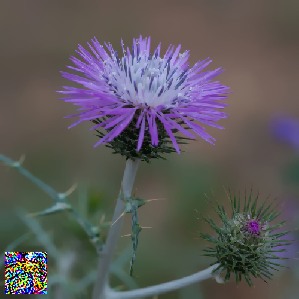}\
    \scriptsize (k) Dragonfly \\ (98\%)
  \end{minipage}
    \begin{minipage}{.135\textwidth}
  	\centering
     \includegraphics[width=\linewidth, keepaspectratio]{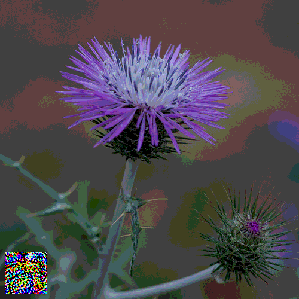}\
    \scriptsize (m) Dragonfly \\ (99\%)
  \end{minipage}
  \vspace{.1cm}
  
    \centering  
  \begin{minipage}{.135\textwidth}
  	\centering
    \includegraphics[width=\linewidth, keepaspectratio]{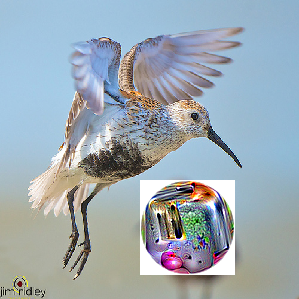}\
    \scriptsize (b) Toaster \\ (94\%)
  \end{minipage}
   \begin{minipage}{.135\textwidth}
   	\centering
    \includegraphics[width=\linewidth,keepaspectratio]{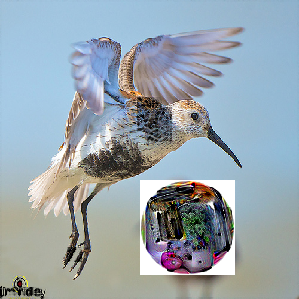}\
     \scriptsize (d) Sandpiper \\ (89\%)
  \end{minipage}
  \begin{minipage}{.135\textwidth}
  	\centering
    \includegraphics[width=\linewidth, keepaspectratio]{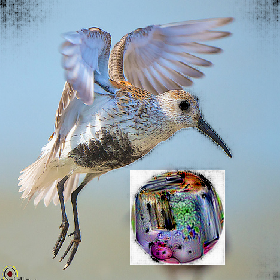}\
    \scriptsize (f) Sandpiper \\ (45\%)
  \end{minipage}
  \begin{minipage}{.135\textwidth}
  	\centering
     \includegraphics[width=\linewidth, keepaspectratio]{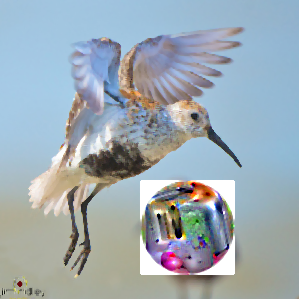}\
    \scriptsize (h) Sandpiper \\ (55\%)
  \end{minipage}
    \begin{minipage}{.135\textwidth}
  	\centering
     \includegraphics[width=\linewidth, keepaspectratio]{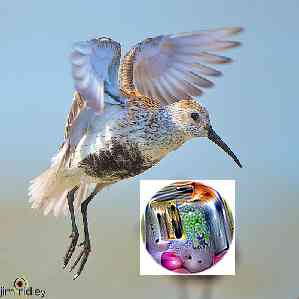}\
    \scriptsize (j) Sandpiper \\ (28\%)
  \end{minipage}
    \begin{minipage}{.135\textwidth}
  	\centering
     \includegraphics[width=\linewidth, keepaspectratio]{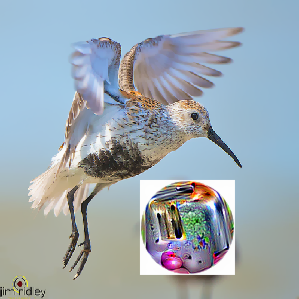}\
    \scriptsize (l) Toaster \\ (90\%)
  \end{minipage}
    \begin{minipage}{.135\textwidth}
  	\centering
     \includegraphics[width=\linewidth, keepaspectratio]{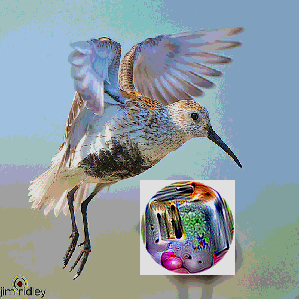}\
    \scriptsize (n) Toaster \\ (92\%)
  \end{minipage}
    
  \end{minipage}
  \vspace{.2cm}
  \caption{Inception v3 confidence score is shown on example images. (a,b) represent  adversarial examples generated by LaVAN and adversarial patch respectively, (c,d) show transformed adversarial images using LGS with lambda equal to 2.3 respectively, (e,f) show transformed adversarial images using DW processing method respectively, (g,h) show transformed adversarial images using median filter with window size 3 respectively, (i,j) show transformed adversarial images using JPEG with quality 30 respectively, (k,l) show transformed adversarial images using TVM with weights equal to 10 respectively, and (m,n) show transformed adversarial images using BR with depth 3.}
\label{fig:all}
\end{figure*}

\subsection{Comparison with Related Defenses}
\label{sec: RW}
In this section, we report comparisons of our approach with other recent defense methods that transform the input sample to successfully reduce the adversarial effect. The compared methods include both global and local techniques. Note that our method processes image locally so it has advantage over other defenses like JPEG, MF, TVM and BR that process image globally. First, we provide a brief description of the competing defenses which will allow us to elaborate further on the performance trends in Tables~\ref{tab:accuracy},~\ref{tab:adv_p} and~\ref{tab:BPDA}.

\subsubsection{Digital Watermarking}
Hayes et.al \cite{hayes2018visible} presented two, non-blind and blind, defense strategies to tackle the challenge of localized attacks \cite{karmon2018lavan,brown2017adversarial}. Non-blind defense considers a scenario, where defender has the knowledge of adversarial mask location. This is unlikely scenario in the context of adversarial attacks because threat is over immediately, once the adversary provides the mask location. Localized attacks have the ability to change the attention of classifier from the original object to adversarial mask. In their blind defense, authors \cite{hayes2018visible} exploited the attention mechanism by first finding the mask location using saliency map and then processing that area before inference. Using saliency map to detect adversarial mask location is the strength of this defense but at the same time its also the weakness of defense because on benign examples, saliency map will give the location of original object and hence processing original object will decrease the performance on clean examples.  Authors \cite{hayes2018visible} reported blind defense performance to protect VGG19 \cite{Simonyan14verydeep} on only 400 randomly selected images with 12\% accuracy drop on clean images. We have tested this defense on imagenet validation set \cite{ILSVRC15} (50k images). This method has the second best accuracy on adversarial examples after LGS but its accuracy on clean examples expectedly dropped by a large margin (22.8\%).  Tables ~\ref{tab:accuracy}, ~\ref{tab:adv_p} and ~\ref{tab:BPDA} summarizes the performance of digital watermarking \cite{hayes2018visible}.

\subsubsection{JPEG Compression}
\cite{dziugaite2016study,das2018shield, shaham2018defending} extensively studied JPEG compression to defend against adversarial attacks. This way high-frequency components are removed that are less important to human vision by using Discrete Cosine Transform (DCT). JPEG performs compression as follows:
\begin{itemize}\setlength{\itemsep}{0em}
\item Convert an RGB image $YC_{b}C_{r}$ color space, where $Y$ and $C_{b}$, $C_{r}$ represent luminance and  chrominance respectively.
\item Down-sample the chrominance channels and apply DCT to $8\times 8$ blocks for each channel.
\item Perform quantization of frequency amplitudes by dividing with a constant and rounding off to the nearest integer.
\end{itemize}
As illustrated in Table~\ref{tab:accuracy}, image quality decreases as the degree of compression increases which in turn decreases accuracy on benign examples. JPEG compression is not very effective against localized attacks, and its defending ability decreases a lot against BPDA. JPEG performance comparison is shown in Tables ~\ref{tab:accuracy},~\ref{tab:adv_p} and ~\ref{tab:BPDA} and Figure~\ref{fig:all}.

\subsubsection{Feature Squeezing}
The main idea of feature squeezing \cite{xu2017feature} is to limit the explorable adversarial space by reducing resolution either by using bit depth reduction or smoothing filters. We found that bit reduction is not effective against localized attacks, however smoothing filter including Gaussian filter, median filter, and bilateral filter reduces localized adversarial effect with reasonable accuracy drop on benign examples. Among smoothing filters, median filter outperforms Gaussian and bilateral filters.
Feature squeezing performance is shown in Tables~\ref{tab:accuracy},~\ref{tab:adv_p} and~\ref{tab:BPDA} and Figure~\ref{fig:all}.

\subsubsection{Total Variance Minimization (TVM)}
Guo \etal \cite{guo2017countering} considered smoothing adversarial images using TVM along with JPEG compression and image quilting. TVM has the ability to measure small variations in the image, and hence it proved effective in removing small perturbations. As illustrated in Table~\ref{tab:accuracy}, TVM becomes ineffective against large concentrated variations introduced by the localized attacks. Further comparisons are shown in Tables~\ref{tab:adv_p} and~\ref{tab:BPDA} and Figure~\ref{fig:all}.

\subsection{Resilience to BPDA}
\label{sec: BPDA}
BPDA \cite{athalye2018obfuscated} is built on the intuition that transformed images by JPEG or TVM should look similar to original images, that is, $\mathcal{T}(x)\approx x$. BPDA approximate gradients for non-differentiable operators with combined forward propagation through operator and DNN while ignoring operator during the backward pass. This strategy allows BPDA to approximate true gradients and thus bypassing the defense. In the traditional attack setting like  Projected Gradient Descent  (PGD) \cite{madry2017towards}, the explorable space available to BPDA is $\mathbb{R}^{n}$ because it can change each pixel in the image. In localized attack setting explorable space reduces to $\mathbb{R}^{m<<n}$ controlled by the mask size. LGS suppresses the high-frequency noise to near zero thereby significantly reducing gradient activity in the estimated mask area and restricting BPDA to bypass defense. However, as it is the case with all defenses, increasing explorable space, i.e., distance limit in PGD attack \cite{madry2017towards} and mask size in the case of localized attack \cite{karmon2018lavan}, protection ability of defense methods decreases. To test performance against BPDA in the localized setting, we randomly selected 1000 examples from Imagenet and attack is optimized against all defenses for the same target, location, mask size and number of iterations. Compared to other defenses methods, LGS significantly reduces the explorable space for localized adversarial attacks within mask size equal to $\sim 2\%$ of the image as discussed in \cite{karmon2018lavan}. In the case of DW \cite{hayes2018visible} defense, we tested BPDA against the proposed input processing given the mask location. Summary of attack success rate against defense methods is presented in Table \ref{tab:BPDA}.

\begin{table}[h]
\centering
\scalebox{0.8}{
\begin{tabularx}{0.6\textwidth}{@{}p{2cm}ccccccc@{}} 
\hline
Defense   & None    &  LGS  & DW & MF & JPEG & TVM & BR \\ \hline
LaVAN with BPDA   & 88\%  &  \bf{18\%} & 25.6\% & 75\%   &  73.30\%     &   78.10\% &   83\%   \\ \hline
\end{tabularx}}
\vspace{0.2cm}
\caption{Attack success rate against Inception v3  with and without defense (lower is better). The size of adversarial noise 42x42 covering $\sim$2\% of image. LGS is used with $\lambda=2.3$, DW in blind scenario, MF with window equal to 3, JPEG compression with quality equal to 30, TVM with weights equal to 10 and BR with depth 3. This hyperparameter choice was made for fair comparison such that the performance on benign examples from ImageNet is approximately the same (first column of Table \ref{tab:accuracy}). Attack is optimized for 1000 randomly selected images for the same target, location and mask size.} 
\label{tab:BPDA}
\end{table}
\section{Discussion and Conclusion}
\label{sec: Conclu}
In this work, we developed a defense against localized adversarial attacks by studying attack properties in gradient domain. Defending against continuously evolving adversarial attacks has proven to be very difficult especially with standalone defenses.  We believe that in critical security applications, a classifier should be replaced by a robust classification system with following main decision stages:
\begin{itemize}[noitemsep,topsep=0pt]
\item \underline{Detection:} given the unlimited distortion space, any image can be converted into an adversarial example that can bypass any defense system with 100\% success rate \cite{athalye2018obfuscated}; however, this also pushes the adversarial example away from the data manifold, and it would be easier to detect rather than removing the perturbation. 
\item \underline{Projection or Transformation:} Adversarial examples within a small distance of original images can be either projected onto the data manifold or transformed to mitigate the adversarial effect.
\item \underline{Classification:} Final stage should be to perform a forward pass through a DNN, whose robustness is increased via adversarial training. 
\end{itemize}
Our method performs a transformation, so it falls into the second stage of robust classification systems. LGS outperforms digital watermarking, JPEG compression, feature squeezing and TVM against localized adversarial attacks with minimal drop in accuracy on benign examples. LGS can be used with a combination of other defense methods, for example, smoothing filters like low pass filter can be applied just on the estimated noisy region to enhance protection for a DNN further. 
{\small
\bibliographystyle{ieee}
\bibliography{egbib}

\begin{thebibliography}{10}\itemsep=-1pt

\bibitem{Akhtar2018ThreatOA}
N.~Akhtar and A.~S. Mian.
\newblock Threat of adversarial attacks on deep learning in computer vision: A
  survey.
\newblock {\em IEEE Access}, 6:14410--14430, 2018.

\bibitem{athalye2018obfuscated}
A.~Athalye, N.~Carlini, and D.~A. Wagner.
\newblock Obfuscated gradients give a false sense of security: Circumventing
  defenses to adversarial examples.
\newblock In {\em International Conference on Machine Learning (ICML)}, 2018.

\bibitem{athalye2017synthesizing}
A.~Athalye, L.~Engstrom, A.~Ilyas, and K.~Kwok.
\newblock Synthesizing robust adversarial examples.
\newblock In {\em International Conference on Machine Learning (ICML)}, 2017.

\bibitem{brown2017adversarial}
T.~B. Brown, D.~Man{\'e}, A.~Roy, M.~Abadi, and J.~Gilmer.
\newblock Adversarial patch.
\newblock In {\em Neural Information Processing Systems (NIPS)}, 2017.

\bibitem{das2018shield}
N.~Das, M.~Shanbhogue, S.-T. Chen, F.~Hohman, S.~Li, L.~Chen, M.~E. Kounavis,
  and D.~H. Chau.
\newblock Shield: Fast, practical defense and vaccination for deep learning
  using jpeg compression.
\newblock In {\em Knowledge Discovery and Data Mining (KDD)}, 2018.

\bibitem{dziugaite2016study}
G.~K. Dziugaite, Z.~Ghahramani, and D.~M. Roy.
\newblock A study of the effect of jpg compression on adversarial images.
\newblock In {\em International Society for Bayesian Analysis (ISBA)}, 2016.

\bibitem{evtimov2017robust}
I.~Evtimov, K.~Eykholt, E.~Fernandes, T.~Kohno, B.~Li, A.~Prakash, A.~Rahmati,
  and D.~Song.
\newblock Robust physical-world attacks on machine learning models.
\newblock In {\em Proceedings of 2018 IEEE Conference on Computer Vision and
  Pattern Recognition (CVPR)}, June 2018.

\bibitem{goodfellow2014explaining}
I.~Goodfellow, J.~Shlens, and C.~Szegedy.
\newblock Explaining and harnessing adversarial examples.
\newblock In {\em International Conference on Learning Representations (ICRL)},
  2015.

\bibitem{guo2017countering}
C.~Guo, M.~Rana, M.~Ciss{\'e}, and L.~van~der Maaten.
\newblock Countering adversarial images using input transformations.
\newblock In {\em International Conference on Learning Representations (ICRL)},
  2017.

\bibitem{hayes2018visible}
J.~Hayes.
\newblock On visible adversarial perturbations \& digital watermarking.
\newblock In {\em Proceedings of the IEEE Conference on Computer Vision and
  Pattern Recognition Workshops}, pages 1597--1604, 2018.

\bibitem{huval2015empirical}
B.~Huval, T.~Wang, S.~Tandon, J.~Kiske, W.~Song, J.~Pazhayampallil,
  M.~Andriluka, P.~Rajpurkar, T.~Migimatsu, R.~Cheng-Yue, et~al.
\newblock An empirical evaluation of deep learning on highway driving.
\newblock {\em arXiv preprint arXiv:1504.01716}, 2015.

\bibitem{karmon2018lavan}
D.~Karmon, D.~Zoran, and Y.~Goldberg.
\newblock Lavan: Localized and visible adversarial noise.
\newblock In {\em International Conference on Machine Learning (ICML)}, 2018.

\bibitem{madry2017towards}
A.~Madry, A.~Makelov, L.~Schmidt, D.~Tsipras, and A.~Vladu.
\newblock Towards deep learning models resistant to adversarial attacks.
\newblock In {\em International Conference on Learning Representations (ICRL)},
  2017.

\bibitem{moosavi2016deepfool}
S.~M. Moosavi~Dezfooli, A.~Fawzi, and P.~Frossard.
\newblock Deepfool: a simple and accurate method to fool deep neural networks.
\newblock In {\em Proceedings of 2016 IEEE Conference on Computer Vision and
  Pattern Recognition (CVPR)}, number EPFL-CONF-218057, 2016.

\bibitem{papernot2016distillation}
N.~Papernot, P.~McDaniel, X.~Wu, S.~Jha, and A.~Swami.
\newblock Distillation as a defense to adversarial perturbations against deep
  neural networks.
\newblock In {\em Security and Privacy (SP), 2016 IEEE Symposium on}, pages
  582--597. IEEE, 2016.

\bibitem{parkhi2015deep}
O.~M. Parkhi, A.~Vedaldi, A.~Zisserman, et~al.
\newblock Deep face recognition.
\newblock In {\em British Machine Vision Conference(BMVC)}, volume~1, page~6,
  2015.

\bibitem{ronen2018microsoft}
R.~Ronen, M.~Radu, C.~Feuerstein, E.~Yom-Tov, and M.~Ahmadi.
\newblock Microsoft malware classification challenge.
\newblock {\em arXiv preprint arXiv:1802.10135}, 2018.

\bibitem{ILSVRC15}
O.~Russakovsky, J.~Deng, H.~Su, J.~Krause, S.~Satheesh, S.~Ma, Z.~Huang,
  A.~Karpathy, A.~Khosla, M.~Bernstein, A.~C. Berg, and L.~Fei-Fei.
\newblock {ImageNet Large Scale Visual Recognition Challenge}.
\newblock {\em International Journal of Computer Vision (IJCV)},
  115(3):211--252, 2015.

\bibitem{shaham2018defending}
U.~Shaham, J.~Garritano, Y.~Yamada, E.~Weinberger, A.~Cloninger, X.~Cheng,
  K.~Stanton, and Y.~Kluger.
\newblock Defending against adversarial images using basis functions
  transformations.
\newblock {\em arXiv preprint arXiv:1803.10840}, 2018.

\bibitem{sharif2016accessorize}
M.~Sharif, S.~Bhagavatula, L.~Bauer, and M.~K. Reiter.
\newblock Accessorize to a crime: Real and stealthy attacks on state-of-the-art
  face recognition.
\newblock In {\em Proceedings of the 2016 ACM SIGSAC Conference on Computer and
  Communications Security}, pages 1528--1540. ACM, 2016.

\bibitem{Simonyan14verydeep}
K.~Simonyan and A.~Zisserman.
\newblock Very deep convolutional networks for large-scale image recognition,
  2014.

\bibitem{su2017one}
J.~Su, D.~V. Vargas, and S.~Kouichi.
\newblock One pixel attack for fooling deep neural networks.
\newblock {\em arXiv preprint arXiv:1710.08864}, 2017.

\bibitem{szegedy2016rethinking}
C.~Szegedy, V.~Vanhoucke, S.~Ioffe, J.~Shlens, and Z.~Wojna.
\newblock Rethinking the inception architecture for computer vision.
\newblock In {\em Proceedings of the IEEE Conference on Computer Vision and
  Pattern Recognition}, pages 2818--2826, 2016.

\bibitem{szegedy2013intriguing}
C.~Szegedy, W.~Zaremba, I.~Sutskever, J.~Bruna, D.~Erhan, I.~Goodfellow, and
  R.~Fergus.
\newblock Intriguing properties of neural networks.
\newblock In {\em International Conference on Learning Representations (ICRL)},
  2014.

\bibitem{tramer2017ensemble}
F.~Tram{\`e}r, A.~Kurakin, N.~Papernot, D.~Boneh, and P.~McDaniel.
\newblock Ensemble adversarial training: Attacks and defenses.
\newblock In {\em International Conference on Learning Representations (ICRL)},
  2018.

\bibitem{xu2017feature}
W.~Xu, D.~Evans, and Y.~Qi.
\newblock Feature squeezing: Detecting adversarial examples in deep neural
  networks.
\newblock {\em arXiv preprint arXiv:1704.01155}, 2017.

\end{thebibliography}
}
\end{document}